\documentclass[lettersize,journal]{IEEEtran}
\usepackage{amsmath,amsfonts}
\usepackage{algorithmic}
\usepackage{algorithm}
\usepackage{array}
\usepackage[caption=false,font=normalsize,labelfont=sf,textfont=sf]{subfig}
\usepackage{booktabs}
\usepackage{multirow}
\usepackage{amsmath}
\usepackage{mathtools}
\usepackage{textcomp}
\usepackage{stfloats}
\usepackage{url}
\usepackage{verbatim}
\usepackage{graphicx}
\usepackage{cite}
\usepackage{caption}
\begin{document}

\title{SocialNav-MoE: A Mixture-of-Experts Vision Language Model \\for Socially Compliant Navigation with Reinforcement Fine-Tuning}

\author{
Tomohito Kawabata$^{1,\ast}$,
Xinyu Zhang$^{1,\ast}$,
Ling Xiao$^{1,\dagger}$~\IEEEmembership{Senior Member,~IEEE}%
\thanks{$^\ast$These authors contributed equally to this work.}%
\thanks{$^\dagger$Corresponding author: \texttt{ling@ist.hokudai.ac.jp}.}%
\thanks{This work was supported by JSPS KAKENHI (Grant No. 24K20787).}%
\thanks{$^1$Graduate School of Information Science and Technology, Hokkaido University, Sapporo, Japan.}%
}



\maketitle

\begin{abstract}
For robots navigating in human-populated environments, safety and social compliance are equally critical, yet prior work has mostly emphasized safety. Socially compliant navigation that accounts for human comfort, social norms, and contextual appropriateness remains underexplored. Vision language models (VLMs) show promise for this task; however, large-scale models incur substantial computational overhead, leading to higher inference latency and energy consumption, which makes them unsuitable for real-time deployment on resource-constrained robotic platforms. To address this issue, we investigate the effectiveness of small VLM and propose SocialNav-MoE, an efficient Mixture-of-Experts vision language model for socially compliant navigation with reinforcement fine-tuning (RFT). We further introduce a semantic similarity reward (SSR) to effectively leverage RFT for enhancing the decision-making capabilities. Additionally, we study the effectiveness of different small language model types (Phi, Qwen, and StableLM), routing strategies, and vision encoders (CLIP vs. SigLIP, frozen vs. fine-tuned). Experiments on the SNEI dataset demonstrate that SocialNav-MoE achieves an excellent balance between navigation accuracy and efficiency. The proposed SSR function is more effective than hard-level and character-level rewards. Source code will be released upon acceptance.
\end{abstract}

\begin{IEEEkeywords}
Socially compliant navigation, efficient vision language model, motion and path planning.
\end{IEEEkeywords}

\section{Introduction}
\IEEEPARstart{W}{ith} the growing deployment of service robots in healthcare, catering, and hospitality, navigation systems are increasingly expected to exhibit human-like, task-oriented behavior~\cite{mirsky2024conflict,xiao2025llm,stratton2024characterizing}. 
Beyond collision avoidance and geometric feasibility, robots must adjust their trajectories and operational states in real time while adhering to social norms and environmental constraints, which requires effective human-robot interaction~\cite{francis2025principles}. 
The core challenge therefore lies in enabling robots to not only react to dynamic environments, but also reason for and model complex human-environment-task relationships in a socially compliant manner.

Existing methods for socially compliant navigation can be broadly categorized into two classes: learning-based approaches and vision language model (VLM)-based approaches. 
Learning-based methods, including reinforcement learning (RL)~\cite{pohland2024stranger} and imitation learning (IL)~\cite{ling2024socialgail}, have demonstrated promising performance in structured settings. 
However, they suffer from several key limitations, including a strong reliance on large-scale, high-quality datasets, sensitivity to hyperparameter configurations, and limited generalization to unseen or complex social scenarios.

In contrast, VLM-based approaches address these challenges by leveraging stronger environmental contextual understanding and commonsense reasoning capabilities~\cite{song2024vlm}. 
Nevertheless, most existing methods rely on large language models (LLMs), whose substantial parameter sizes introduce high inference latency and hinder real-time deployment on robots~\cite{elnoor2025vlm,mavrogiannis2023core}. 
Moreover, as these models are typically trained on general-purpose image-text data, they often struggle with high-level decision-making and fine-grained social reasoning required for socially compliant navigation~\cite{guo2023images}.

To address the issues mentioned above, we propose SocialNav-MoE, an efficient VLM for socially compliant navigation with reinforcement fine-tuning. SocialNav-MoE adopts a multimodal small language model (SLM) built on a sparse mixture-of-experts (MoE) architecture as the baseline, and novelly leverages reinforcement fine-tuning (RFT) strategy to enhance decision-making capabilities. Specifically, we introduce a novel semantic similarity reward (SSR) function. SSR effectively mitigates the sparse reward problem inherent in the hard-level and character-level rewards commonly used in existing RFT methods. Finally, we systematically investigate key design factors, including the choice of SLM backbones, the number of experts (1–4), routing strategies, the selection and training schemes of vision encoders.

Overall, the contributions of this work can be summarized as follows:
\begin{itemize}
    \item We propose SocialNav-MoE, an efficient VLM for socially compliant navigation with RFT. SocialNav-MoE achieves an excellent balance between navigation accuracy and efficiency, demonstrating that a well-designed combination of MoE structure and RFT significantly improves decision-making for socially compliant navigation under scarce data settings.
    \item We introduce a novel SSR function, which is substantially more effective than the hard-level and character-level rewards commonly employed in RFT.
    \item We systematically analyze the impact of SLM backbones, expert configurations, and vision encoder designs, providing practical guidelines for socially compliant navigations.
\end{itemize}

\section{Related Work}

\subsection{Socially Compliant Robot Navigation}
Socially compliant navigation goes beyond collision avoidance, requiring robots to follow implicit social norms such as interpersonal distance, right-of-way, and culturally preferred motion patterns~\cite{karnan2022socially}. These norms are context-dependent and rarely explicit, making robust modeling in dynamic human environments challenging. Most prior work adopts learning-based approaches, including demonstration-driven motion learning~\cite{sun2021motion} and reinforcement learning in simulation~\cite{liang2021crowd}. Although effective in controlled settings, these methods rely on large-scale data or realistic simulators and often struggle to generalize to unseen social scenarios. Recently, VLMs have enabled socially aware navigation through stronger contextual reasoning, supporting high-level action generation~\cite{payandeh2024social}, trajectory evaluation~\cite{narasimhan2024olivia}, and human motion prediction~\cite{song2024vlm}. Large-scale datasets such as SCAND~\cite{karnan2022socially} and MuSoHu~\cite{nguyen2023toward} further promote this direction. However, most VLM-based approaches rely on computationally expensive LLMs, and efficient socially compliant navigation remains underexplored.

\subsection{VLMs for Navigation}
VLM have recently advanced autonomous navigation by enabling high-level reasoning in complex, context-rich environments~\cite{yin2024sg,raj2024rethinking}. By jointly modeling vision and language, VLMs provide a unified framework for perception, planning, and decision-making. Early VLM-based navigation approaches encode visual observations or scene semantics into structured language prompts, allowing LLMs to generate navigation decisions using pretrained knowledge. Representative examples include NAVGPT~\cite{zhou2024navgpt}, NAVGPT2~\cite{zhou2024navgpt2}, and LLM-Advisor~\cite{xiao2025llm}. Subsequent works extended this paradigm to zero-shot navigation, such as VLTNet with Tree-of-Thought reasoning for language-driven object navigation~\cite{wen2025zero} and UniGoal, which adopts unified graph representations for general zero-shot navigation~\cite{yin2025unigoal}. More recent methods move beyond purely language-driven planning. For instance, NoMaD~\cite{sridhar2024nomad} introduces a diffusion-based unified policy for both exploration and goal-conditioned navigation in unseen environments. However, most VLM-based navigation systems rely on large-scale models, leading to high computational costs and limited applicability to real-time, resource-constrained robotic platforms.

\subsection{Reinforcement Fine-Tuning}
Post-training of VLMs has traditionally relied on SFT with annotated multimodal data to improve task-specific performance. 
In contrast, RFT has recently shown strong ability to enhance the reasoning capabilities of LLMs, especially for mathematical problem solving and code generation~\cite{tan2025reason, liu2025visual, luo2025gson}. 
Representative approaches such as Direct Preference Optimization (DPO)~\cite{rafailov2023direct} and reinforcement learning with verifiable rewards (RLVR) methods like Group Relative Policy Optimization (GRPO)~\cite{shao2024deepseekmath}, together with the success of models such as DeepSeek~\cite{guo2025deepseek}, demonstrate that reinforcement learning alone can substantially improve reasoning performance. 

Despite these advances, the adoption of RFT in VLMs remains limited. Existing studies~\cite{liu2025nav,qi2025vln} mainly target final-stage language-level objectives (e.g., hallucination mitigation or preference alignment), rather than in-loop decision-making for embodied or interactive tasks. 
Consequently, the feasibility of applying modern RFT paradigms to socially compliant navigation has not been systematically examined. 
To address this gap, we systematically explore RFT for socially compliant navigation and introduce a novel semantic similarity reward function to enhance decision-making capabilities.

\section{METHOD}

\subsection{Overall Architecture}
\begin{figure*}[t]
  \centering
  \includegraphics[width=0.8\textwidth]{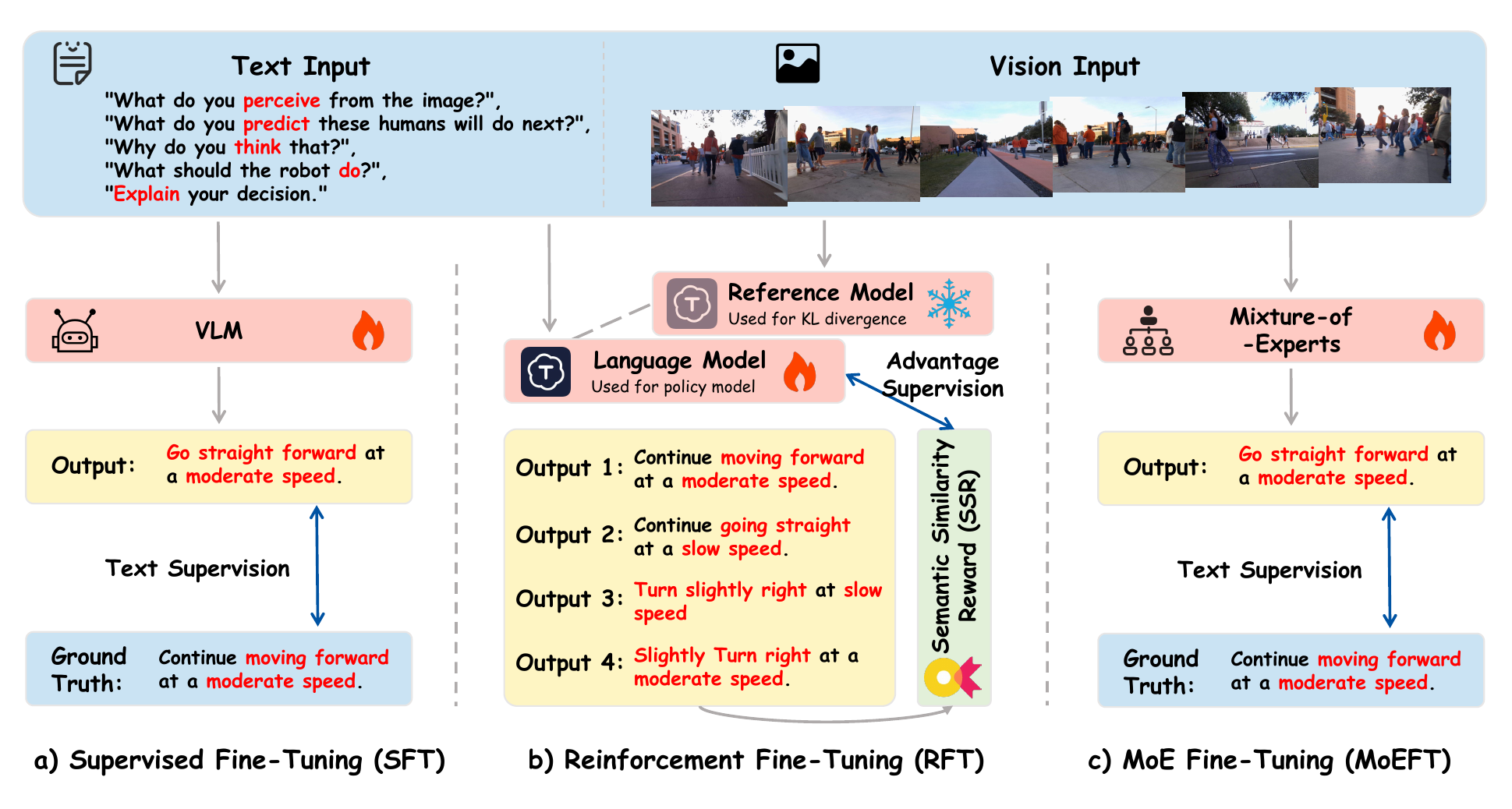}
  \caption{The overall architecture of SocialNav-MoE. SocialNav-MoE consists of three training stages: supervised fine-tuning, reinforcement fine-tuning, and MoE fine-tuning.}
  \label{fig:MoEllavanav}
\end{figure*}
Fig.~\ref{fig:MoEllavanav} illustrates the overall pipeline of SocialNav-MoE. Given image scene descriptions from the SNEI dataset showing Fig.~\ref{fig:dataset}, we construct prompt–response pairs as inputs to the VLM for model fine-tuning. Overall, the model is trained in three stages:
\begin{figure}[t]
  \centering
  \includegraphics[width=\linewidth]{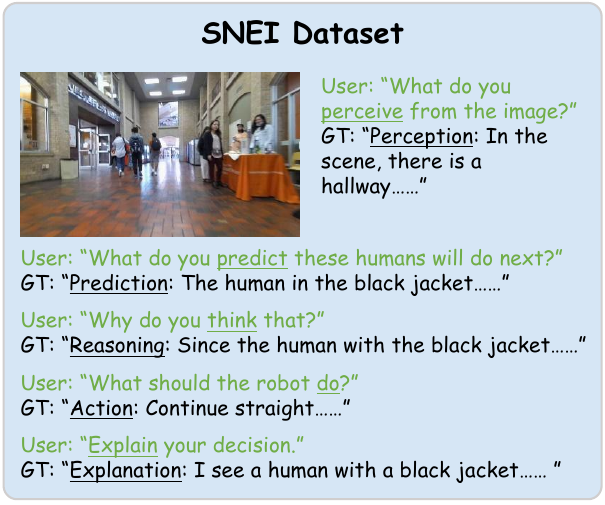}
  \caption{SNEI dataset example.}
  \label{fig:dataset}
\end{figure}

\noindent(a) Supervised Fine-Tuning (SFT). 
The SocialNav-MoE is trained via supervised language modeling with ground-truth navigation actions to generate navigation behaviors from multimodal inputs.

\noindent(b) Reinforcement Fine-Tuning (RFT). 
The SLM used in SocialNav-MoE is further optimized to align its outputs with both ground-truth actions and human subjective preferences. We propose a novel SSR function to effectively leverage Group Sequence Policy Optimization (GSPO)~\cite{zheng2025group} for RFT.

\noindent(c) MoE Fine-Tuning (MoEFT). 
The mixture-of-experts (MoE) structure is optimized using the RFT-enhanced SLM. 
Multi-turn conversational fine-tuning is applied to produce stable and coherent motion actions.


\subsection{Supervised Fine-Tuning}

\noindent\textbf{Step I (Image–Text Alignment).}
Given visual tokens $V_p$ extracted from the input image by the vision encoder and text tokens $T_q$ obtained from the corresponding caption sequence, where $p$ and $q$ index the visual and textual token sets and $P$ and $Q$ denote the corresponding index sets, we optimize image–text alignment via a captioning objective:

\begin{equation}
\mathcal{L}_{\text{I}} 
= - \sum_{n=1}^{N} \log \pi_\theta \left( T_q \mid V_p \right), 
\quad (p \in P,\; q \in Q).
\end{equation}

\noindent\textbf{Step II (SNEI Fine-Tuning).}  
For multi-turn $t$ conversation, we form multimodal context
\[
\alpha_t = [V_p, (p \in P), \ T_q, (q \in Q)].
\]
Supervised fine-tuning is employed to guide the VLM in learning the mapping from multi-turn $t$ conversation to motion action for socially compliant navigation tasks.

\subsection{Reinforcement Fine-Tuning}
\noindent\textbf{Recap of GSPO.} GSPO~\cite{zheng2025group} adopts a simplified optimization strategy that assigns uniform weights to all tokens within a response. By treating each response as a whole rather than applying token-level weights, GSPO provides more stable optimization for MoE-based models. The optimization objective of GSPO is given below:
\begin{equation}
\begin{multlined}
J_{\mathrm{GSPO}}(\theta)=\mathbb{E}\Bigg[\frac{1}{G} \sum_{i=1}^{G} \min \Big(s_{i}(\theta) A_{i}, \\
\operatorname{clip}(s_{i}(\theta), 1-\epsilon, 1+\epsilon) A_{i}\Big)-\beta D_{\mathrm{KL}}(\pi_{\theta} \| \pi_{\mathrm{ref}})\Bigg],
\end{multlined}
\end{equation}
where \(G\) denotes the number of sampled responses within a group. \(\beta\) is the coefficient of Kullback-Leibler (KL) divergence. \(\epsilon\) represents coefficient of clip. 
The importance sampling ratio \(s_i(\theta)\) is defined as
\begin{equation}
s_{i}(\theta)=
\left(
\frac{\pi_{\theta}(y_{i} \mid x)}
{\pi_{\theta_{\text{old}}}(y_{i} \mid x)}
\right)^{\frac{1}{|y_{i}|}},
\end{equation}
where \(y_i\) denotes the \(i\)-th generated response for the input query \(x\), and \(|y_i|\) is the response length. 
GSPO applies length normalization to reduce variance and keep the importance ratio within a stable numerical range, which is particularly notable for training stability in sparse Mixture-of-Experts (MoE) structure. The advantage function is shown in the following
\begin{equation}
A_i =\frac{R(x, y_i) - \mathrm{mean}(\{R(x, y_i)\}_{i=1}^{G})}
{\mathrm{std}(\{R(x, y_i)\}_{i=1}^{G})},
\end{equation}
where \(A_i\) represents the relative advantage of the \(i\)-th response, and \(R(x, y_i)\) denotes its corresponding reward. 
This formulation encourages the model to favor responses with higher relative rewards within the group.
To constrain the magnitude of policy updates and improve training stability, a KL divergence regularization term is introduced
\begin{equation}
D_{\mathrm{KL}}(\pi_{\theta} \| \pi_{\mathrm{ref}}) =
\frac{\pi_{\mathrm{ref}}(y_i \mid x)}{\pi_{\theta}(y_i \mid x)}
- \log \frac{\pi_{\mathrm{ref}}(y_i \mid x)}{\pi_{\theta}(y_i \mid x)} - 1.
\end{equation}
this KL term maintains proximity between the updated policy \(\pi_{\theta}\) and the reference policy \(\pi_{\mathrm{ref}}\), thereby improving both training stability and sample efficiency.

However, due to the task-dependent nature of $R(x, y_i)$, GSPO cannot be directly applied to our task. 
As a result, designing an effective reward function to leverage GSPO for socially compliant navigation remains an open problem.

\noindent\textbf{Proposed SSR.}
Existing reward functions commonly used in prior work include hard-level reward~\cite{liu2025nav} and character-level reward.
Eq.~\ref{hard-level} formulates the hard-level reward, where a reward is assigned only when the generated action exactly matches a predefined target.
In contrast, Eq.~\ref{charac-level} defines the character-level reward, which evaluates performance based on lexical overlap at the character level.

\begin{equation}
\label{hard-level}
R_{hard}(x, y_i) =
\begin{cases}
1, & \text{if } y_i = \text{ground truth}, \\
0, & \text{otherwise}.
\end{cases}
\end{equation}

\begin{equation}
\label{charac-level}
R_{character}(x, y_i) =
\frac{|C_{y_i} \cap C_{ref}|}{\max(1, |C_{ref}|)},
\end{equation}
where \(C_{y_i} \cap C_{ref}\) represents the intersection of the \(i\)-th response and ground truth on the character set, and \(\max(1, |C_{ref}|)\) denotes the total number of unique characters in the ground truth.

However, both reward formulations are inherently rigid and therefore unsuitable for socially compliant navigation. Hard-level reward enforce exact action matching and provide feedback only at discrete target states, while character-level rewards depend on strict character-to-character alignment at the surface level. As a result, both fail to accommodate the semantic variability and contextual flexibility required in social navigation scenarios.

Therefore, we propose semantic similarity reward (SSR), which leverages BERTScore-F1. By exploiting contextual embeddings from a pre-trained BERT backbone, SSR enables token alignment based on semantic correspondence rather than surface-level string overlap.
Specifically, given a generated output \(y\) and a ground truth \(g\), a pretrained BERT encoder~\cite{devlin2019bert} produces token embeddings \(\{e_y^1,\ldots,e_y^n\}\) and \(\{e_g^1,\ldots,e_g^m\}\), with $m$ and $n$ as the number of tokens in the response and ground truth, respectively. 
The cosine similarity between token pairs forms a similarity matrix \(\mathbf{S} \in \mathbb{R}^{m \times n}\), where \(\mathbf{S}_{jk} = \mathrm{cosine}(e_y^j, e_g^k)\). 
Based on \(\mathbf{S}\), recall \(R\) and precision \(P\) are computed as
\begin{equation}
R = \frac{1}{m} \sum_{k=1}^{m} \max_{1 \leq j \leq n} S_{jk}, \quad
P = \frac{1}{n} \sum_{j=1}^{n} \max_{1 \leq k \leq m} S_{jk},
\end{equation} 
The BERTScore-F1 is then obtained as the harmonic mean of precision and recall:
\begin{equation}
\text{BERTScore-F1} = \frac{2PR}{P + R},
\end{equation}
using precision alone for reward tends to encourage short and overly conservative responses, while recall alone often leads to verbose and redundant outputs. 
By balancing precision and recall, BERTScore-F1 provides a more informative reward signal for guiding high-level decision-making in socially compliant navigation. Therefore, the semantic similarity reward function is as follows:
\begin{equation}
R_{SSR}(x, y_i) = \text{BERTScore-F1}(x, y_i)
\end{equation}

\subsection{MoE Fine-Tuning} 
The MoE architecture~\cite{eigen2013learning} employs a router to score experts and enable sparse activation. We utilize top-$k$ routing, where only the $k$ highest scoring experts (out of $K$ total experts) are evaluated. The router maps an input token $\alpha$ to a length-$K$ score vector:
\begin{equation}
W(\alpha)_{i}=\frac{e^{f(\alpha)_{i}}}{\sum_{j=1}^{K} e^{f(\alpha)_{j}}},
\end{equation}
where $f(\alpha)_i$ is the $i$-th logit from the router’s linear layer. The MoE output is
\begin{equation}
MoE(\alpha)=\sum_{i=1}^{k} W(\alpha)_{i}F(\alpha)_{i},
\end{equation}
where $F(\alpha)_{i}$ is the output of the $i$-th selected expert.
In this stage, we adopt multi-turn conversational fine-tuning and inference to train the MoE architecture with dynamic expert routing. 
Each conversation turn is routed to the most relevant expert, enhancing contextual understanding. 
Multi-turn interactions also improve contextual coherence, enabling the MoE to produce stable and socially compliant behaviors.

\section{EXPERIMENTS AND DISCUSSIONS}

\subsection{Implementation Details}

\noindent\textbf{Experimental Setup.} We adopt SNEI~\cite{payandeh2024social} as the experimental dataset, which consists of 325 images, with 265 images used for training and 60 for testing. 
Non-geometric transformation based data augmentation is applied to the original training set to simulate variations in lighting and weather conditions, resulting in an augmented training set of 530 images. 
SocialNav-MoE is trained on the augmented training set and evaluated on the 60 testing set.

SigLIP~\cite{zhai2023sigmoid} is used as the vision encoder, and Phi-2-2.7B~\cite{javaheripi2023phi} is adopted as the SLM. 
To balance representation capacity and training stability, we employ an alternating feed-forward network (FFN) design, in which half of the transformer layers are replaced with sparse MoE layers to enhance expressive power, while the remaining layers retain standard FFNs to ensure stable optimization. 
This design serves as the baseline architecture of SocialNav-MoE.

Training is performed in multiple stages. 
First, the model is initialized from LLaVA-1.5-58K and trained for one iteration cycle with a learning rate of \(1\times10^{-3}\). 
Next, SFT is conducted on the augmented SNEI dataset for 20 iteration cycles using a reduced learning rate of \(2\times10^{-5}\). 
RFT is then applied for 3 epochs with a learning rate of \(2\times10^{-6}\), during which both GRPO+SSR (Ours) and GSPO+SSR (Ours) are evaluated with a group size of 8. 
Finally, MoE fine-tuning (MoEFT) is performed on the augmented SNEI dataset for 20 epochs using a learning rate of \(2\times10^{-6}\).

\noindent\textbf{Evaluation Metrics.} We utilize the total number of model parameters and frames per second (FPS) to evaluate the model's size and inference efficiency respectively. To preserve consistency with the training input format and maximize context utilization, we employ a multi-turn conversation during inference. Given that action instructions are the primary evaluation criteria of socially compliant behavior, we only measure inference time exclusively for the action instructions generation phase across these conversation for FPS. For performance evaluation, we focus on next-step action prediction from SocialNav-MoE and validate their plausibility using multiple semantic similarity metrics: BERTScore for token-level semantic alignment, SBERT for sentence-level cosine similarity in embedding space, and sentence mover’s similarity (SMS) for holistic semantic correspondence across sentences.

\subsection{Main Results}
The comparisons are summarized in Tables~\ref{tab:effic} and~\ref{tab:metrics}. We also compare with two off-the-shelf VLMs with strong generalization ability (GPT-4o, Claude), which are considered to have certain social skills~\cite{thapa2024gpt}. Specifically, we directly employ these two pre-trained models to conduct inference experiments on the SNEI dataset. As we can see, the total number of parameters for GPT-4o and Claude are about 200B~\cite{abacha2024medec} and 175B~\cite{abacha2024medec}, respectively, and FPS are 0.212 and 0.087. In contrast, our method has a total parameter count of 5.74B and FPS of 1.709. The parameter quantities only account for 2.9\% and 3.3\% of GPT-4o and Claude, and FPS is 8.1 times and 19.6 times higher than GPT-4o and Claude, respectively. Considering SMS, which captures broader contextual meaning and complements the token-level BERTScore and cosine similarity-based SBERT, the following text compares the results of SMS. Our method achieved the best performance in all metrics, with SMS being 46.5\% and 32.1\% higher than GPT-4o and Claude, respectively. Compared to using the original data, the augmented data shows improvements in all metrics.
\begin{table}[t!]
\centering
\caption{Efficiency comparisons of different models. The best performances are highlighted in bold.}
\label{tab:effic}
\resizebox{\columnwidth}{!}{%
\begin{tabular}{lcccc}
\toprule
Model & SLMs & Vision Encoder & Parameter & FPS \\
\midrule
GPT-4o & -& - & 200 B & 0.212 \\
Claude &- & - & 175 B & 0.087 \\
SocialNav-MoE & Phi-2-2.7B~\cite{javaheripi2023phi} & SigLIP~\cite{zhai2023sigmoid} & \textbf{5.74} B & \textbf{1.709} \\
\bottomrule
\end{tabular}
}
\vspace{0.5em}
\centering
\caption{Accuracy comparisons of different models. The best performances are highlighted in bold.}
\label{tab:metrics}
\resizebox{\columnwidth}{!}{%
\begin{tabular}{lcccccc}
\toprule
Model & Training dataset & BERTScore-P & BERTScore-R & BERTScore-F1 & SBERT-cos & SMS \\
\midrule
GPT-4o  & - & 0.076 & 0.443 & 0.254 & 0.557 & 0.376 \\
Claude  & - & -0.233 & 0.387 & 0.059 & 0.509 & 0.417 \\
\multirow{2}{*}{SocialNav-MoE} & Original dataset & 0.445 & 0.410 & 0.427 & 0.594 & 0.489 \\
 & Augmented dataset & \textbf{0.520} & \textbf{0.492} & \textbf{0.506} & \textbf{0.671} & \textbf{0.551} \\
\bottomrule
\end{tabular}
}
\end{table}

\begin{figure*}[t!]
  \centering
  \includegraphics[width=\linewidth]{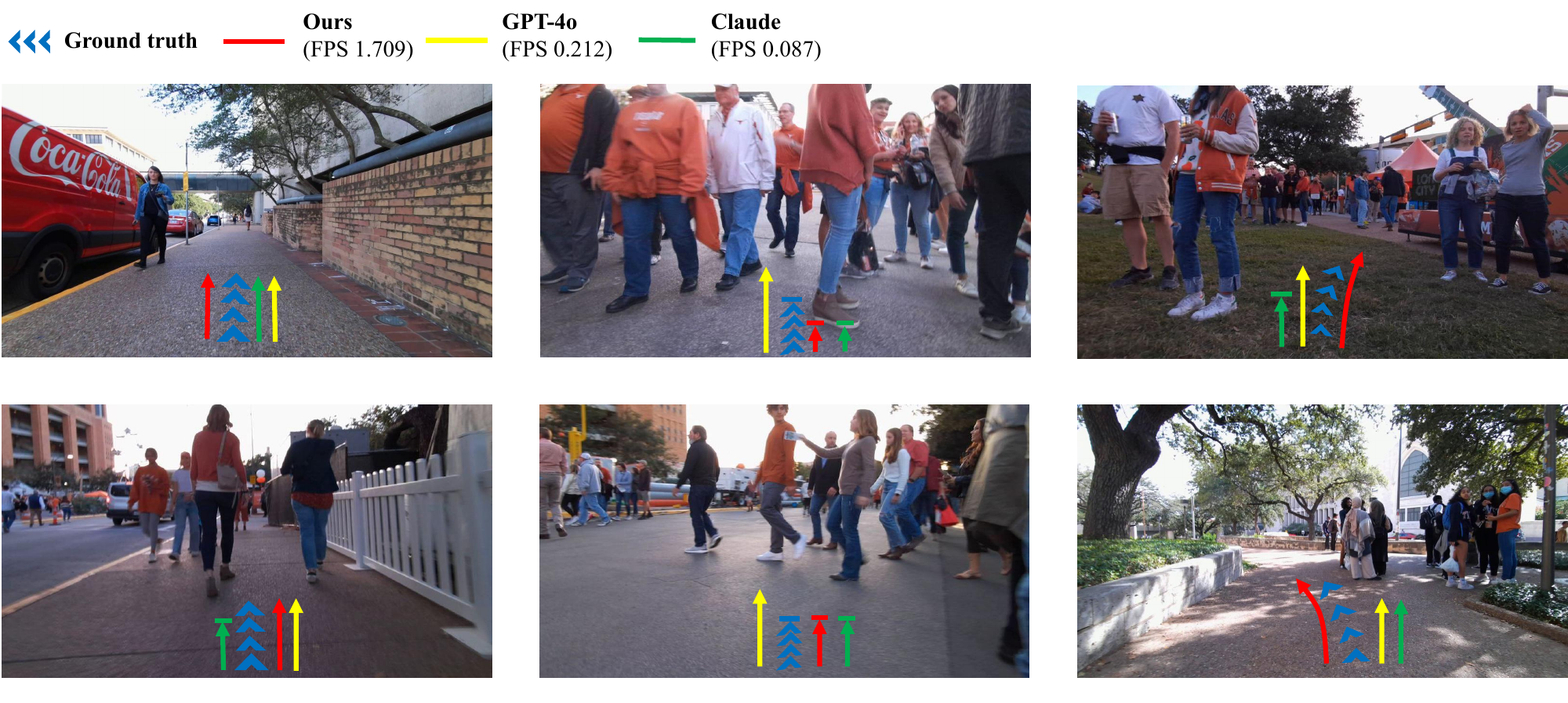}
  \caption{Visualization of the model outputs. The blue arrow indicates the ground truth, the red arrow denotes our method, the yellow arrow represents GPT-4o, and the green arrow corresponds to Claude. GPT-4o and Claude often fail to produce socially compliant motions. For example, in the top-right image, GPT-4o predicts a forward action and Claude predicts a stop action, where a slight right turn would be more appropriate. In the central top and bottom images, GPT-4o also suggests moving forward when stopping is preferable. These cases highlight the advantages of our approach in maintaining fluid and socially aware navigation.}
  \label{fig:visual}
\end{figure*}

Visualization results are provided in Fig.~\ref{fig:visual}. As illustrated, the robot adheres to socially compliant navigation behaviors: maintaining speed when the path is clear, executing slow turns to bypass stationary crowds, and stopping (indicated by a horizontal line) for moving obstructions. Qualitatively, our method (red arrow) aligns closely with the ground truth (blue arrow). In contrast, large model baselines like Claude (green) and GPT-4o (yellow) frequently deviate from social norms, often failing to execute necessary bypass maneuvers or stops.

\subsection{Discussions}
\noindent\textbf{Varying the SLMs.} 
We evaluate the results when varying the SLMs. As shown in Table~\ref{tab:varying-SLMs}, when the vision encoder is fixed to SigLIP~\cite{zhai2023sigmoid}, Phi achieves the best performance across all metrics among the three SLMs. Phi's SBERT-cos metrics are 0.8\% and 13.3\% higher than Qwen and StableLM, respectively. Therefore we adopted the Phi model for subsequent experiments.
\begin{table}[t!]
\centering
\caption{Comparison of varying SLMs.}
\label{tab:varying-SLMs}
\resizebox{\columnwidth}{!}{%
\begin{tabular}{lccccc}
\toprule
SLMs & BERTScore-P & BERTScore-R & BERTScore-F1 & SBERT-cos & SMS \\
\midrule
Phi-2-2.7B~\cite{javaheripi2023phi} & \textbf{0.486} & \textbf{0.466} & \textbf{0.476} & \textbf{0.642} & \textbf{0.523} \\
Qwen-1.8B~\cite{bai2023qwen} & 0.486 & 0.456 & 0.472 & 0.636 & 0.520 \\
StableLM-1.6B~\cite{bellagente2024stable} & 0.419 & 0.378 & 0.399 & 0.566 & 0.461 \\
\bottomrule
\end{tabular}
}
\end{table}

\noindent\textbf{Examining the Expert Routing.} 
We examine the effect of expert routing on model performance. 
As shown in Table~\ref{tab:topk}, when top-$k$=1 (i.e., only the expert with the highest activation weight is selected), increasing the number of experts consistently improves performance. 
Specifically, SMS increases by approximately $10.6\%$ as the number of experts increasing from 1 to 4, suggesting that a larger experts pool enables better specialization and more effective multimodal reasoning.

A similar trend is observed when top-$k$=2, where performance improves as the number of experts increases from 1 to 3. 
However, further increasing the number of experts to 4 leads to a performance degradation. 
We attribute this drop to data scarcity: activating more experts introduces additional noise due to feature divergence among experts, resulting in conflicting signals during output aggregation. 
Overall, comparisons across different $k$ values and expert configurations consistently support this observation.

\begin{table*}[t!]
\centering
\caption{Comparison of different $k$ values in top-$k$ routing across different numbers of experts.}
\label{tab:topk}
\begin{tabular}{c|c|ccccc}
\hline
Experts & Top-$k$ & BERTScore-P & BERTScore-R & BERTScore-F1 & SBERT-cos & SMS \\
\hline
1 & 1 & 0.418 & 0.414 & 0.416 & 0.595 & 0.473 \\
\hline
\multirow{2}{*}{2}
 & 1 & 0.463 & 0.447 & 0.455 & 0.631 & 0.519 \\
 & 2 & 0.423 & 0.418 & 0.421 & 0.597 & 0.479 \\
\hline
\multirow{3}{*}{3}
 & 1 & 0.460 & 0.434 & 0.447 & 0.615 & 0.517 \\
 & 2 & 0.442 & 0.431 & 0.437 & 0.610 & 0.494 \\
 & 3 & 0.402 & 0.399 & 0.400 & 0.583 & 0.478 \\
\hline
\multirow{4}{*}{4}
 & 1 & \textbf{0.486} & \textbf{0.466} & \textbf{0.476} & \textbf{0.642} & \textbf{0.523} \\
 & 2 & 0.414 & 0.406 & 0.410 & 0.588 & 0.475 \\
 & 3 & 0.407 & 0.402 & 0.404 & 0.587 & 0.473 \\
 & 4 & 0.398 & 0.388 & 0.393 & 0.581 & 0.471 \\
\hline
\end{tabular}
\end{table*}

\noindent\textbf{Varying the Vision Encoders.} We compare the impact of different vision encoder types (Table~\ref{tab:encoder-clip-siglip}) and training paradigms (Table~\ref{tab:encoder}) on model performance, with the number of experts fixed at 4 and top-$k$ set to 1. 
As shown in Table~\ref{tab:encoder-clip-siglip}, SigLIP~\cite{zhai2023sigmoid} slightly outperforms CLIP~\cite{radford2021learning}, yielding an SMS improvement of approximately 1.8\%. 
We further examine different training paradigms for the vision encoder based on SigLIP.
Under limited data settings, freezing the vision encoder achieves better performance than full fine-tuning, improving SMS by 1.6\%.
Moreover, freezing the encoder significantly reduces the number of trainable parameters.
Considering both performance gains and training efficiency, we adopt the frozen vision encoder configuration in our final model.

\begin{table}[t!]
\centering
\caption{Varying vision encoders in SocialNav-MoE.}
\label{tab:encoder-clip-siglip}
\resizebox{\columnwidth}{!}{%
\begin{tabular}{c|ccccc}
\hline
Vision encoder & BERTScore-P & BERTScore-R & BERTScore-F1 & SBERT-cos & SMS \\
\hline
CLIP~\cite{radford2021learning}   & 0.473 & 0.457 & 0.465 & 0.638 & 0.514 \\
SigLIP~\cite{zhai2023sigmoid} & \textbf{0.486} & \textbf{0.466} & \textbf{0.476} & \textbf{0.642} & \textbf{0.523} \\
\hline
\end{tabular}
}
\vspace{0.5em}
\centering
\caption{Varying training paradigm for vision encoder.}
\label{tab:encoder}
\resizebox{\columnwidth}{!}{%
\begin{tabular}{c|ccccc}
\hline
Training paradigm & BERTScore-P & BERTScore-R & BERTScore-F1 & SBERT-cos & SMS \\
\hline
Frozen  & 0.486 & 0.466 & 0.476 & \textbf{0.642} & \textbf{0.523} \\
Trained & \textbf{0.487} & \textbf{0.468} & \textbf{0.478} & 0.641 & 0.515 \\
\hline
\end{tabular}
}
\end{table}

\begin{figure}[t!]
  \centering
  \includegraphics[width=\linewidth]{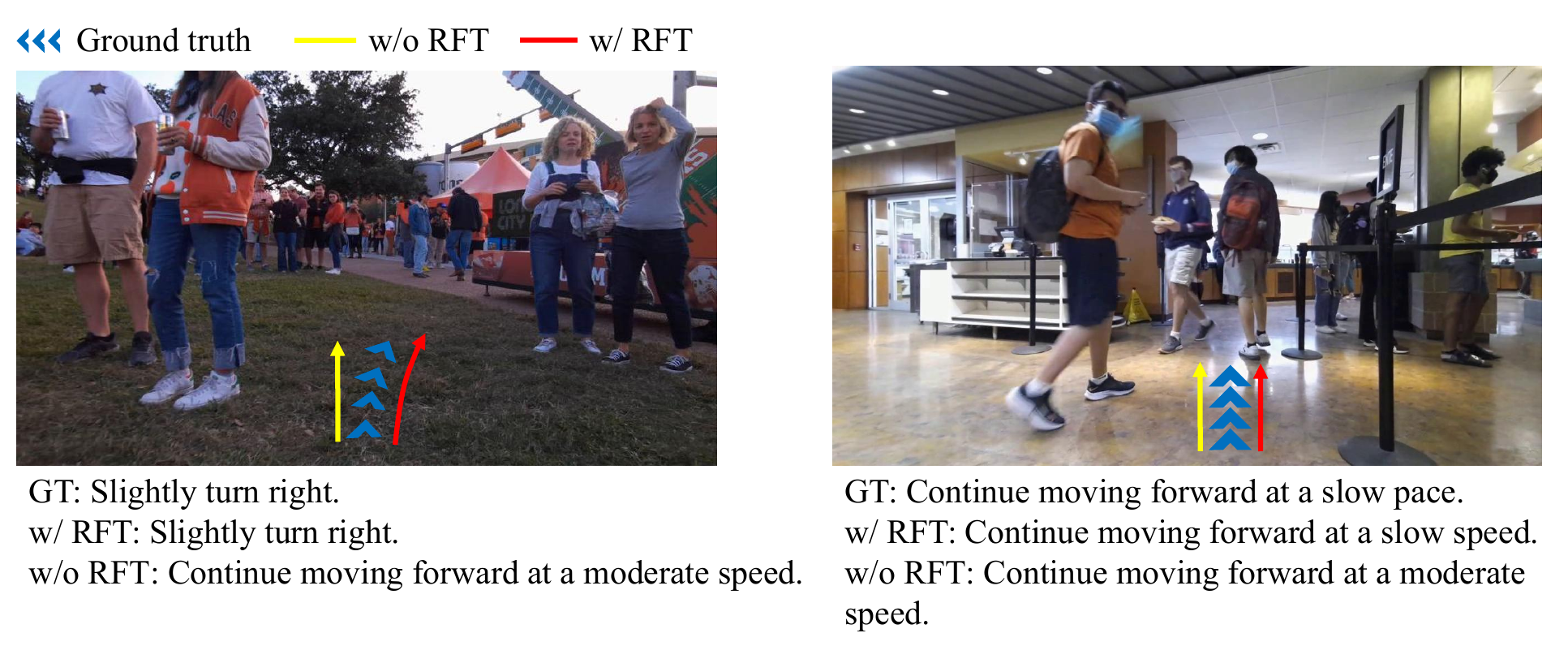}
  \caption{Visualization of action outputs with and without RFT. 
Arrows indicate trajectories: blue for ground truth (GT), red for SocialNav-MoE with RFT, and yellow for SocialNav-MoE without RFT. 
\textbf{Left:} With RFT, the model correctly aligns with the GT by predicting a slight right turn, whereas without RFT it fails to deviate and proceeds forward at a moderate speed. 
\textbf{Right:} Although both models predict a straight trajectory, the RFT-enhanced model accurately matches the GT’s low speed, while the model without RFT incorrectly predicts a higher speed. 
These examples demonstrate that RFT substantially improves socially compliant navigation.
}
  \label{fig:rft}
\end{figure}

\noindent\textbf{Comparison with Typical RFT Methods.}
We compare the performance of different RFT methods using 4 experts with top-$k$=1 in Table~\ref{tab:rft}. 
Note that GRPO and GSPO cannot be directly applied to this task; therefore, we adapt them by incorporating the proposed SSR. 
Compared with DPO, GRPO+SSR yields moderate performance gains, suggesting that leveraging group relative advantages provides a more informative learning signal than pairwise direct preference optimization. 
GSPO+SSR further improves performance, achieving a 5.4\% increase in SMS. 
These results indicate that sequence-level group preference optimization in GSPO enables more stable and sample-efficient policy updates, ultimately leading to more reasonable socially compliant behavior.
Fig.~\ref{fig:rft} further visualizes the effects of RFT. 
With RFT, SocialNav-MoE not only predicts the correct action but also accurately estimates motion speed, resulting in more socially compliant navigation compared to SocialNav-MoE without RFT.

\begin{table*}[t]
\centering
\begin{minipage}{0.48\textwidth}
\centering
\captionof{table}{Comparison of RFT methods.}
\label{tab:rft}
\resizebox{\columnwidth}{!}{%
\begin{tabular}{c|ccccc}
\hline
RFT & BERTScore-P & BERTScore-R & BERTScore-F1 & SBERT-cos & SMS \\
\hline
DPO   & 0.481 & 0.463 & 0.472 & 0.649 & 0.527 \\
GRPO+SSR (Ours)  & 0.499 & 0.467 & 0.483 & 0.655 & 0.531 \\
GSPO+SSR (Ours) & \textbf{0.520} & \textbf{0.492} & \textbf{0.506} & \textbf{0.671} & \textbf{0.551} \\
\hline
\end{tabular}
}
\end{minipage}
\hfill
\begin{minipage}{0.48\textwidth}
\centering
\captionof{table}{Comparison of different reward types.}
\label{tab:reward}
\resizebox{\columnwidth}{!}{%
\begin{tabular}{c|ccccc}
\hline
Reward type & BERTScore-P & BERTScore-R & BERTScore-F1 & SBERT-cos & SMS \\
\hline
Hard level       & 0.488 & 0.466 & 0.477 & 0.645 & 0.532 \\
Character Level  & 0.453 & 0.435 & 0.444 & 0.627 & 0.510 \\
SSR              & \textbf{0.520} & \textbf{0.492} & \textbf{0.506} & \textbf{0.671} & \textbf{0.551} \\
\hline
\end{tabular}
}
\end{minipage}

\vspace{0.2em}
\centering
\caption{different train/inference settings of SocialNav-MoE.}
\label{tab:rft-sft-gspo}
\begin{tabular}{c|c|c|ccccc}
\hline
SocialNav-MoE & Train & Inference & BERTScore-P & BERTScore-R & BERTScore-F1 & SBERT-cos & SMS \\
\hline
\multirow{2}{*}{w/o RFT}
 & Multi-turn & Multi-turn & 0.486 & 0.466 & 0.476 & 0.642 & 0.523 \\
 & Single-turn & Single-turn & 0.466 & 0.453 & 0.460 & 0.624 & 0.510 \\
\hline
\multirow{2}{*}{w/ RFT}
 & Multi-turn & Multi-turn & \textbf{0.520} & \textbf{0.492} & \textbf{0.506} & \textbf{0.671} & \textbf{0.551} \\
 & Single-turn & Single-turn & 0.468 & 0.457 & 0.463 & 0.637 & 0.514 \\
\hline
\end{tabular}
\end{table*}


\begin{figure}[t!]
  \centering
  \includegraphics[width=\linewidth]{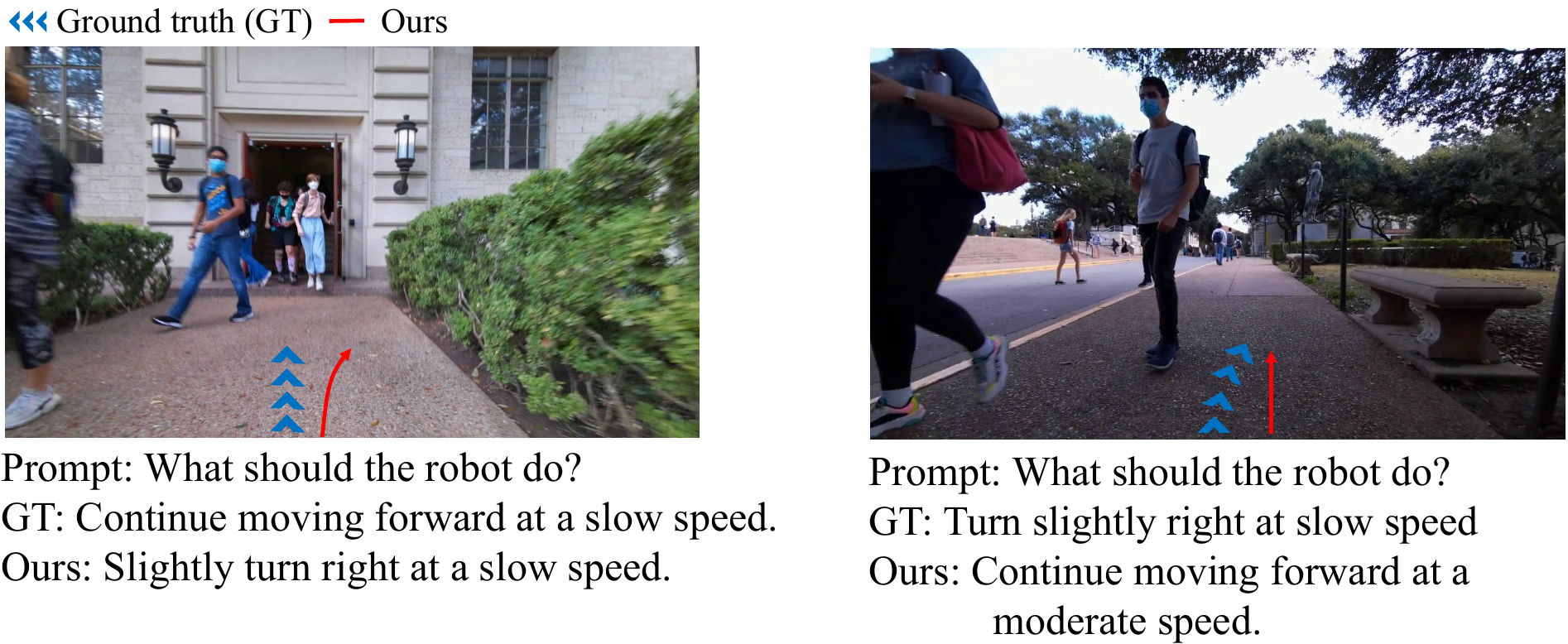}
  \caption{Visualization of failure cases. The blue arrow indicates the ground truth (GT), the red arrow denotes our method. \textbf{Left:} GT suggests continuing straight at a slow speed, whereas our model predicts a slight right turn; \textbf{Right:} GT favors a slight right turn, while our model continues forward at a moderate speed. Both decisions are socially acceptable, reflecting the inherent ambiguity in defining universally valid navigation norms. These cases illustrate that deviations from GT do not always correspond to clear mistakes, but rather to alternative interpretations of socially compliant behavior.}
  \label{fig:fail}
\end{figure}

Table~\ref{tab:reward} compares different reward types. 
The hard-level reward, defined in Eq.~\ref{hard-level}, assigns supervision only when the generated response exactly matches the ground truth, resulting in sparse reward signals that fail to reflect partial correctness. 
The character-level reward, defined in Eq.~\ref{charac-level}, measures lexical overlap at the character level by computing the recall of ground-truth characters in the generated response. 
Although it provides denser feedback, this reward is highly sensitive to noise, which introduces misleading guidance and degrades performance.
In contrast, the proposed SSR leverages BERTScore to evaluate semantic similarity between generated responses and ground truth, effectively bridging the gap between exact matching and semantic correctness. 
As a result, models trained with SSR achieve superior performance, outperforming the hard-level and character-level rewards by 4.0\% and 7.0\% in SMS, respectively. 
These results highlight the importance of semantic-aware rewards in mitigating reward sparsity and noise, thereby enabling more effective RFT training.

\noindent\textbf{Varying the Train/Inference Data Settings.}
Table~\ref{tab:rft-sft-gspo} reports an ablation study examining the effects of interaction setting (single-turn vs. multi-turn) and RFT on model performance. 
Across all metrics, the multi-turn setting consistently outperforms the single-turn baseline, regardless of whether RFT is applied. 
This indicates that multi-turn interactions enable the model to better leverage contextual information and capture long-range dependencies, leading to more socially compliant navigation behaviors. 
In addition, incorporating RFT yields substantial performance gains, demonstrating its effectiveness in refining the policy space and aligning navigation decisions more closely with human social preferences.

\noindent\textbf{Failure Analysis and Future Works.} While our method demonstrates competitive performance, the lack of universally accepted social norms poses challenges for both model learning and evaluation. 
As illustrated in Fig.~\ref{fig:fail}, it is often ambiguous whether the ground truth or the model’s output better reflects appropriate social behavior. 
Addressing this ambiguity remains an important direction for future work. 
In addition, we plan to deploy our model on real robotic platforms and further explore lightweight social navigation models and practical deployment solutions for real-world applications.

\section{CONCLUSION}

In this work, we present SocialNav-MoE, an efficient framework for socially compliant robot navigation that achieves a strong balance between navigation performance and inference efficiency. 
To enhance decision-making capabilities, we introduce a novel SSR function, which enables effective RFT for socially compliant navigation. 
We further conduct a systematic evaluation of SLMs, expert configurations, routing strategies, and vision encoders, yielding several key insights.
First, scaling the number of experts to four with a Top-$k$=1 routing strategy significantly improves performance, resulting in a 10.6\% increase in SMS. 
This configuration outperforms Top-$k$=2 by providing a better trade-off between expert diversity and inference efficiency. 
Second, freezing the vision encoder yields a 1.6\% improvement in SMS, indicating that fixed visual representations are preferable in data-scarce regimes to mitigate overfitting. 
Third, the choice of vision encoder has a tangible impact on performance, with SigLIP outperforming CLIP by 1.8\% in SMS, demonstrating better alignment with semantic similarity objectives. 
Finally, incorporating an intermediate RFT stage substantially enhances decision-making capability: specifically, GSPO combined with the proposed SSR achieves an additional 3.4\% improvement in SMS.

\bibliographystyle{IEEEtran}
\bibliography{refs}

\end{document}